\documentclass[11pt,a4paper]{article}

\usepackage[utf8]{inputenc}
\usepackage[T1]{fontenc}
\usepackage{amsmath,amssymb}
\usepackage{graphicx}
\usepackage{booktabs}
\usepackage[hidelinks]{hyperref}
\usepackage{xcolor}
\usepackage{listings}
\usepackage{float}
\usepackage{caption}
\usepackage{subcaption}
\usepackage[margin=1in]{geometry}
\usepackage{enumitem}
\usepackage{fancyhdr}
\usepackage{pgfplots}
\pgfplotsset{compat=1.18}
\usepackage{tikz}
\usepackage{url}
\usepackage{natbib}

\definecolor{codebg}{rgb}{0.95,0.95,0.97}
\definecolor{codegreen}{rgb}{0.25,0.50,0.35}
\definecolor{codegray}{rgb}{0.50,0.50,0.50}
\definecolor{codepurple}{rgb}{0.58,0.00,0.83}
\definecolor{codeblue}{rgb}{0.10,0.10,0.60}

\lstdefinestyle{jsonStyle}{
    backgroundcolor=\color{codebg},
    basicstyle=\ttfamily\small,
    breaklines=true,
    captionpos=b,
    commentstyle=\color{codegreen},
    keywordstyle=\color{codeblue}\bfseries,
    stringstyle=\color{codepurple},
    showstringspaces=false,
    frame=single,
    framerule=0.4pt,
    rulecolor=\color{codegray},
    xleftmargin=6pt,
    xrightmargin=6pt,
    aboveskip=8pt,
    belowskip=8pt,
}

\lstdefinestyle{pythonStyle}{
    backgroundcolor=\color{codebg},
    basicstyle=\ttfamily\small,
    breaklines=true,
    captionpos=b,
    commentstyle=\color{codegreen},
    keywordstyle=\color{codeblue}\bfseries,
    stringstyle=\color{codepurple},
    showstringspaces=false,
    frame=single,
    framerule=0.4pt,
    rulecolor=\color{codegray},
    xleftmargin=6pt,
    xrightmargin=6pt,
    language=Python,
}

\title{%
  \textbf{JTON: A Token-Efficient JSON Superset with\\Zen Grid Tabular Encoding for Large Language Models}
}

\author{
  Gowthamkumar Nandakishore \\
  \texttt{github.com/gowthamkumar-nandakishore/JTON}
}

\date{March 2026}

\begin{document}
\maketitle

\begin{abstract}
When LLMs process structured data, the serialization format directly
affects cost and context utilization. Standard JSON wastes tokens repeating
key names in every row of a tabular array--overhead that scales linearly with
row count. This paper presents \textbf{JTON} (JSON Tabular Object Notation),
a strict JSON superset whose main idea, \textbf{Zen Grid}, factors column
headers into a single row and encodes values with semicolons, preserving
JSON's type system while cutting redundancy. Across seven real-world domains,
Zen Grid reduces token counts by \textbf{15--60\%} versus JSON compact
(28.5\% average; 32\% with \texttt{bare\_strings}).

Comprehension tests on \textbf{10~LLMs} show a net +0.3\,pp accuracy gain
over JSON: four models improve, three hold steady, and three dip slightly.
Generation tests on \textbf{12~LLMs} yield 100\% syntactic validity in both
few-shot and zero-shot settings. A Rust/PyO3 reference implementation adds
SIMD-accelerated parsing at 1.4$\times$ the speed of Python's
\texttt{json} module. Code, a 683-vector test suite, and all experimental
data are publicly available.
\end{abstract}

\vspace{0.5em}
\noindent\textbf{Keywords:} JTON, Zen Grid, token-efficient serialization, LLM-native data formats, JSON superset, structured data for LLMs, SIMD parsing

\section{Introduction}
\label{sec:intro}

Large language models now sit at the center of most programmatic data
pipelines, consuming structured prompts and emitting tool-call
responses~\citep{Brown2020GPT3}. In this setting, every token of embedded
data counts against context limits and billing meters, so the choice of
serialization format is no longer just a convenience--it is an architectural
decision with direct cost implications. Applications that feed tabular data
into LLM prompts (database results, API payloads, analytics tables) feel this
most acutely: the format dictates how much data fits in a context window and
what each call costs.

The formats in widespread use today--JSON, CSV, YAML, Markdown tables--were
all designed before LLMs existed. They optimize for human readability or
machine parsing, not for \emph{token efficiency}. The LLM era, we argue,
demands a new class of serialization formats: ones designed to minimize
token counts while staying readable by both people and language models.

JSON~\citep{RFC8259} is the default interchange format for LLM-facing APIs,
but it prioritizes self-description at the expense of compactness.
Consider a typical pattern: an array of objects sharing a schema:

\begin{lstlisting}[style=jsonStyle, caption={JSON compact: 116 characters{,} \texttildelow42 tokens}]
[{"id":1,"name":"Alice","score":95},{"id":2,"name":"Bob","score":87},
 {"id":3,"name":"Carol","score":92}]
\end{lstlisting}

\noindent
Every row repeats the key names \texttt{"id"}, \texttt{"name"}, and
\texttt{"score"}, along with structural characters (\texttt{\{}, \texttt{\}},
\texttt{:}, quotes). For a table with $k$ columns and $n$ rows, standard JSON
repeats the $k$ keys $n$ times, producing $O(n \cdot k)$ overhead in key
tokens alone.

This redundancy is a property of the \emph{format}, not the data. Tabular
data has a natural layout in which headers appear once and rows hold only
values:

\begin{lstlisting}[style=jsonStyle, caption={JTON Zen Grid: 67 characters{,} \texttildelow28 tokens (33\% fewer)}]
[3: id, name, score; 1, "Alice", 95; 2, "Bob", 87; 3, "Carol", 92 ]
\end{lstlisting}

\noindent
The \texttt{[3:} prefix declares three data rows; the header row (\texttt{id,
name, score}) appears once; semicolons delimit rows. The entire structure
round-trips through \texttt{jton.loads()} to produce the same Python list of
dicts as the JSON original.

This paper presents \textbf{JTON} (JSON Tabular Object Notation), a strict
JSON superset, and its central contribution \textbf{Zen Grid}--a compact,
LLM-friendly tabular encoding. Our contributions:

\begin{enumerate}[leftmargin=*,itemsep=2pt]
  \item \textbf{Zen Grid format}: A tabular syntax that cuts token counts by
    15--60\% on real-world data versus JSON compact (up to 61\% with
    \texttt{bare\_strings}), scaling with row count and key length
    (\S\ref{sec:token}, \S\ref{sec:realworld}).
  \item \textbf{Format comparison}: Head-to-head benchmarks against CSV,
    Markdown, and YAML showing Zen Grid as the most token-efficient format
    that keeps JSON's type system (\S\ref{sec:alternatives}).
  \item \textbf{LLM comprehension study}: Ten models, five task types, seven
    domains--net result +0.3\,pp over JSON compact at 32\% fewer tokens
    (\S\ref{sec:llm}, \S\ref{sec:realworld}).
  \item \textbf{LLM generation study}: Twelve models, 100\% validity in
    both few-shot and zero-shot prompting (\S\ref{sec:generation}).
  \item \textbf{SIMD-accelerated implementation}: A Rust+PyO3 parser using
    AVX2/AVX-512 structural scanning, achieving 1.2--1.6$\times$ speedup over
    Python's \texttt{json} module (\S\ref{sec:perf}).
  \item \textbf{Open-source release}: Library, 683-vector test suite, and all
    experimental code at
    \url{https://github.com/gowthamkumar-nandakishore/JTON}.
\end{enumerate}

\section{Related Work}
\label{sec:related}

\paragraph{Data interchange formats.}
JSON~\citep{RFC8259} is ubiquitous for web APIs and LLM tool outputs.
YAML~\citep{YAML} trades compactness for readability (20--80\% more tokens
than JSON); TOML~\citep{TOML} targets configuration files. Binary formats
like MessagePack~\citep{MessagePack}, CBOR~\citep{RFC7049}, and Protocol
Buffers~\citep{Protobuf} achieve excellent compactness but cannot be embedded
directly in text prompts. CSV matches Zen Grid on tabular compactness but
sacrifices type information--there is no way to distinguish a string from a
number, null, or boolean--and nested data is out of reach. Zen Grid keeps
JSON's type system while gaining the tabular layout.

\paragraph{Token-aware data formats.}
TOON~\citep{TOON} uses a table-oriented notation to save about 19\% of
tokens, but it is not JSON-compatible and needs a custom parser.
JTON sits in a different part of the design space: roughly 20\% savings, but
any standard JSON parser can consume JTON output that does not use Zen Grid
extensions.

\paragraph{Prompt compression.}
LLMLingua~\citep{LLMLingua} and related systems compress natural language by
stripping low-information tokens. That work is orthogonal to ours: they
operate on prose; Zen Grid operates on structured data.

\paragraph{High-performance JSON parsers.}
simdjson~\citep{Langdale2019} pioneered SIMD structural scanning via VPSHUFB
nibble classification, pushing throughput past 2~GB/s.
orjson~\citep{orjson} brings similar ideas to a Python-facing Rust library
($\sim$730~MB/s), and yyjson~\citep{yyjson} offers a C alternative. JTON's
parser borrows the nibble-classifier approach from simdjson and string-caching
patterns from orjson, extending the grammar to handle Zen Grid syntax and
JTON extensions.

\paragraph{Alternative tabular representations.}
CSV strips key repetition but throws away types: there is no way to tell a
blank cell from \texttt{null}, and nested values simply cannot be expressed.
Markdown tables pad cells for visual alignment, inflating token counts.
YAML block-style list-of-maps still repeats every key and adds indentation
overhead. OpenAI's structured-output modes define schemas separately from
data but still serialize the payload as plain JSON, so the key-repetition
problem remains. Section~\ref{sec:alternatives} gives a quantitative
comparison.

\section{Format Specification}
\label{sec:format}

JTON is a strict superset of JSON: every valid JSON document parses as valid
JTON. Three extensions are added; Zen Grid is the one that matters most.

\subsection{JSON Extensions}

\paragraph{Unquoted keys.}
Object keys matching the pattern \texttt{[a-zA-Z\_][a-zA-Z0-9\_]*} may omit
quotes: \lstinline[style=jsonStyle]|{name: "Alice", age: 30}|.

\paragraph{Comments.}
Both \texttt{//} line comments and \texttt{/* */} block comments are allowed
anywhere whitespace is permitted.

\paragraph{Special numbers.}
The literals \texttt{Infinity}, \texttt{-Infinity}, and \texttt{NaN} are
recognized as IEEE~754 special values, mapping to Python's
\texttt{float("inf")} and \texttt{float("nan")}.

\subsection{Zen Grid Tables}
\label{sec:zengrid}

Zen Grid is the heart of JTON: a compact way to encode arrays of objects
that share a common key set.

\paragraph{Syntax.}
A Zen Grid table is enclosed in \texttt{[N:\,\ldots\,]} where $N$ is the
(optional) row count, followed by a header row (comma-separated column names)
and data rows (semicolon-delimited):

\begin{lstlisting}[style=jsonStyle]
[N: <header1>, <header2>, ..., <headerK>;
    <row1_val1>, <row1_val2>, ..., <row1_valK>;
    <row2_val1>, <row2_val2>, ..., <row2_valK> ]
\end{lstlisting}

\paragraph{Grammar.}
In extended BNF:
\begin{lstlisting}[basicstyle=\ttfamily\small,frame=single,backgroundcolor=\color{codebg}]
zen_grid     = "[" [row_count] ":" headers (";" row)* "]"
row_count    = non_negative_integer
headers      = header ("," header)*
header       = json_string | identifier
row          = cell ("," cell)*
cell         = json_value | identifier | <empty>
identifier   = [a-zA-Z_][a-zA-Z0-9_]*
\end{lstlisting}

\noindent
Both \texttt{[:} (no row count) and \texttt{[N:} (with row count) are valid
prefixes. The row count, when present, serves as a structural hint that aids
both human readers and LLM generation.

\paragraph{Semantics.}
A Zen Grid table \texttt{[2: h1, h2; v1, v2; v3, v4~]} is semantically
equivalent to the JSON array
\texttt{[\{"h1":v1,"h2":v2\},\{"h1":v3,"h2":v4\}]}.
Missing cells (fewer values than headers) are interpreted as \texttt{null}.

\paragraph{Serialization options.}
JTON provides additional options for maximum token efficiency:
\begin{itemize}[itemsep=1pt]
  \item \texttt{bare\_strings=True}: Unquoted identifier-like string values
    in cells (e.g., \texttt{Alice} instead of \texttt{"Alice"}), saving 5--10\%
    additional tokens on string-heavy data.
  \item \texttt{implicit\_null=True}: Empty cells represent \texttt{null}
    instead of the literal \texttt{null}, saving tokens on sparse tables.
\end{itemize}

\paragraph{Detection heuristic.}
When serializing, a Python list is converted to Zen Grid if: (a) it contains
$\geq$2 elements, (b) all elements are dicts, and (c) $\geq$70\% of elements
share the same key set as the first element.

\paragraph{Token savings analysis.}
For a table with $k$ columns and $n$ rows, where $\bar{t}_h$ is the average
header token count:

\begin{equation}
\Delta T = (n - 1) \cdot k \cdot (\bar{t}_h + t_{\text{struct}})
\label{eq:savings}
\end{equation}

\noindent
where $t_{\text{struct}} \approx 3$ accounts for the quotes, colon, and
braces per key-value in JSON. Savings grow linearly with $n$ and $k$.

\section{Token Efficiency Evaluation}
\label{sec:token}

We measure token counts using the \texttt{o200k\_base} tokenizer
(tiktoken~\citep{tiktoken}), used by GPT-4o and GPT-5 class models.

\subsection{Methodology}

We generate synthetic tabular datasets with controlled parameters:
\begin{itemize}[itemsep=1pt]
  \item \textbf{Employee records}: 4 columns (id, name, dept, salary),
    string-heavy
  \item \textbf{Product inventory}: 5 columns (sku, name, category, price,
    stock), mixed types
  \item \textbf{Server metrics}: 4 columns (timestamp, cpu, memory, requests),
    number-heavy
\end{itemize}

Each dataset is serialized as (1) JSON pretty-printed, (2) JSON compact
(\texttt{separators=(",",":")}), (3) JTON Zen Grid, and (4) JTON Zen Grid
with \texttt{bare\_strings=True}. Token counts are measured at row counts from
5 to 1{,}000.

\subsection{Results}

\begin{table}[H]
\centering
\caption{Token counts by format and dataset size (employee records, 4 columns).}
\label{tab:token_scaling}
\begin{tabular}{@{}rrrrrr@{}}
\toprule
\textbf{Rows} & \textbf{JSON Pretty} & \textbf{JSON Compact} & \textbf{Zen Grid} & \textbf{Zen Bare} & \textbf{$\Delta$ vs Compact} \\
\midrule
5     & 168     & 88      & 75     & 64    & $-$14.8\% \\
10    & 335     & 175     & 142    & 119   & $-$18.9\% \\
25    & 836     & 436     & 343    & 284   & $-$21.3\% \\
50    & 1{,}671 & 871     & 678    & 559   & $-$22.2\% \\
100   & 3{,}342 & 1{,}742 & 1{,}349& 1{,}109& $-$22.6\% \\
250   & 8{,}351 & 4{,}351 & 3{,}358& 2{,}759& $-$22.8\% \\
500   & 16{,}702& 8{,}702 & 6{,}709& 5{,}509& $-$22.9\% \\
1{,}000& 33{,}403& 17{,}403& 13{,}410& 11{,}010& $-$22.9\% \\
\bottomrule
\end{tabular}
\end{table}

\begin{figure}[H]
\centering
\includegraphics[width=0.85\columnwidth]{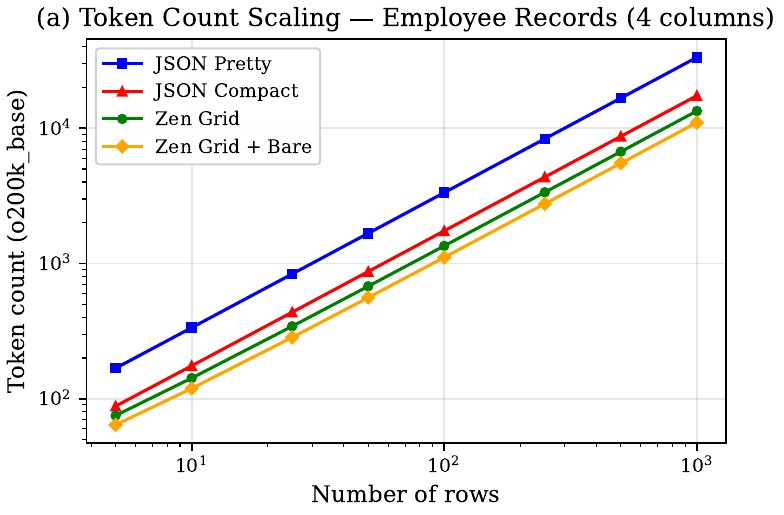}
\caption{Token count scaling for employee records (4 columns,
\texttt{o200k\_base} tokenizer). Zen Grid savings increase with row count,
converging to $\sim$23\% at scale.}
\label{fig:token_scaling}
\end{figure}

\begin{table}[H]
\centering
\caption{Token savings across different data shapes (100 and 500 rows).}
\label{tab:cross_dataset}
\begin{tabular}{@{}lrrrrr@{}}
\toprule
\textbf{Dataset} & \textbf{Rows} & \textbf{Cols} & \textbf{JSON Tokens} & \textbf{Zen Tokens} & \textbf{Savings} \\
\midrule
Employees   & 100 & 4 & 1{,}742 & 1{,}349 & $-$22.6\% \\
Employees   & 500 & 4 & 8{,}702 & 6{,}709 & $-$22.9\% \\
Products    & 100 & 5 & 2{,}764 & 2{,}273 & $-$17.8\% \\
Products    & 500 & 5 & 13{,}838& 11{,}347& $-$18.0\% \\
Metrics     & 100 & 4 & 2{,}694 & 2{,}301 & $-$14.6\% \\
Metrics     & 500 & 4 & 13{,}472& 11{,}479& $-$14.8\% \\
\bottomrule
\end{tabular}
\end{table}

\paragraph{Discussion.}
Savings range from 15\% on number-heavy data (metrics with long timestamps)
up to 23\% on string-heavy records (short, repeated keys). Convergence is
fast: by 50 rows the per-row amortization of the header is negligible.
Turning on \texttt{bare\_strings} adds roughly 14\,pp on top
(e.g., 22.9\% $\rightarrow$ 36.7\% at 1{,}000 rows). These numbers line up
with the theoretical model in Eq.~\ref{eq:savings}, which predicts savings
proportional to key length and row count.

\begin{figure}[H]
\centering
\includegraphics[width=0.85\columnwidth]{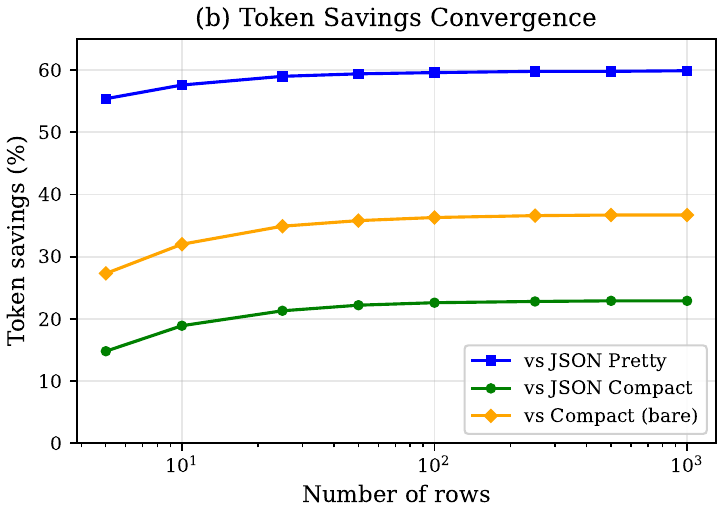}
\caption{Token savings (\%) versus JSON formats converge rapidly with row
count, reaching 23\% versus compact and 59\% versus pretty-printed by
1{,}000 rows.}
\label{fig:savings_convergence}
\end{figure}

\begin{figure}[H]
\centering
\includegraphics[width=0.9\columnwidth]{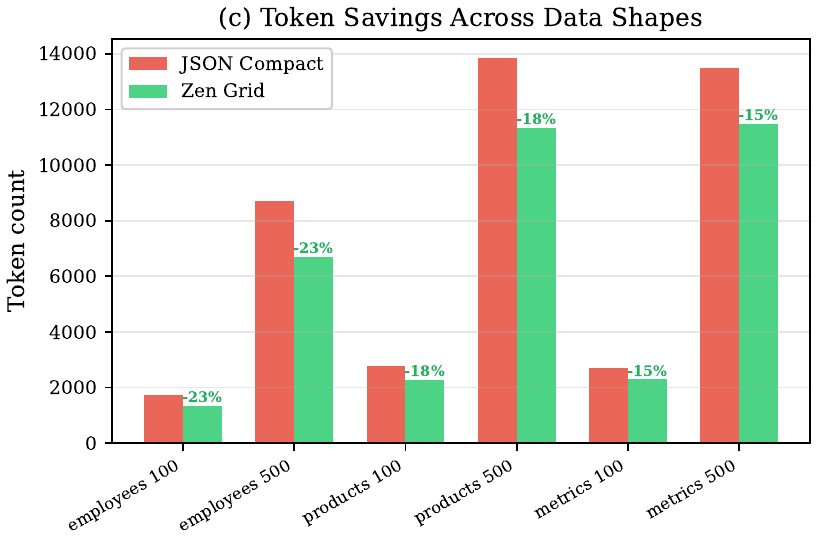}
\caption{Zen Grid token savings across different data shapes. Savings range
from 15\% (number-heavy metrics) to 23\% (string-heavy employees).}
\label{fig:cross_dataset}
\end{figure}

\section{Real-World Data Evaluation}
\label{sec:realworld}

Synthetic benchmarks are useful for controlled comparisons, but real-world
data is messier--longer keys, mixed types, occasional nulls. To check that
Zen Grid holds up, we evaluate on seven real-world datasets across different
domains.

\subsection{Datasets}

We use three external datasets from standard JSON benchmarks--%
\textbf{Twitter} (20 tweets from the Twitter Search API, 7 fields including
user handles, tweet text with Unicode, and engagement metrics),
\textbf{GitHub} (20 events from the GitHub Events API, 6 fields including
nested actor/repo references), and \textbf{CITM} (20 ticketing performance
records from the CitM catalog, 6 fields with large numeric IDs and
timestamps)--plus four generated real-world datasets: \textbf{Financial}
(20 stock trades with nulls and mixed types), \textbf{Weather} (25 station
readings with 10 columns), \textbf{Healthcare} (20 patient lab results with
boolean flags), and \textbf{Logistics} (20 shipping records with nested
status/priority).

\subsection{Token Efficiency on Real-World Data}

\begin{table}[H]
\centering
\caption{Token counts and savings across seven real-world datasets (20--25
rows each, \texttt{o200k\_base} tokenizer, Zen Grid with \texttt{bare\_strings=True}).}
\label{tab:real_token}
\begin{tabular}{@{}lrrrrr@{}}
\toprule
\textbf{Dataset} & \textbf{Rows} & \textbf{JSON Compact} & \textbf{Zen Bare} & \textbf{$\Delta$ Compact} & \textbf{$\Delta$ Pretty} \\
\midrule
Twitter (API)     & 20 & 3{,}673 & 1{,}422 & $-$61.3\% & $-$65.9\% \\
GitHub (Events)   & 20 & 968     & 780     & $-$19.4\% & $-$43.8\% \\
CITM (Ticketing)  & 20 & 871     & 612     & $-$29.7\% & $-$52.6\% \\
Financial (Trades)& 20 & 1{,}358 & 1{,}041 & $-$23.3\% & $-$48.5\% \\
Weather (Stations)& 25 & 1{,}911 & 1{,}316 & $-$31.1\% & $-$52.3\% \\
Healthcare (Lab)  & 20 & 1{,}349 & 936     & $-$30.6\% & $-$53.3\% \\
Logistics (Ship.) & 20 & 1{,}333 & 936     & $-$29.8\% & $-$53.0\% \\
\midrule
\textbf{Average}  &    &         &         & \textbf{$-$32.2\%} & \textbf{$-$52.8\%} \\
\bottomrule
\end{tabular}
\end{table}

\noindent
With \texttt{bare\_strings=True}, real-world data yields substantially higher
savings than synthetic data (32.2\% average vs.~compact, compared to the
15--23\% on synthetic sets). The reason is straightforward: production schemas
use longer, more descriptive key names--\texttt{retweet\_count},
\texttt{favorite\_count}, \texttt{ordering\_physician}--and their repetition
eats up the token budget. Bare strings then save additional tokens on
identifier-like values. Twitter hits 61.3\% because its 7-column tweet objects
have especially long keys and abundant string values. Against pretty-printed
JSON, savings average 52.8\%. Without \texttt{bare\_strings}, standard Zen
Grid still manages 15--60\% (28.5\% average).

\begin{figure}[H]
\centering
\includegraphics[width=0.85\columnwidth]{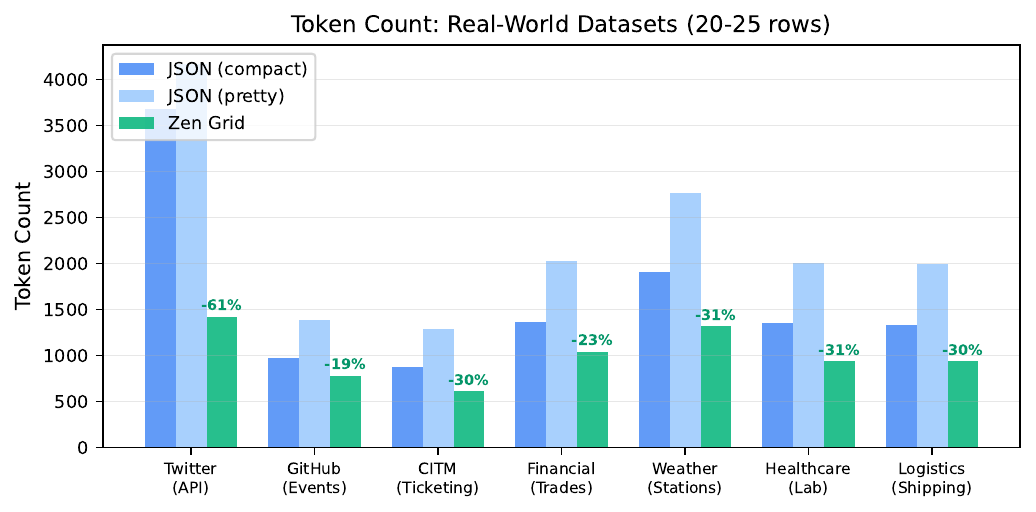}
\caption{Token counts across seven real-world datasets. Zen Grid (green)
consistently uses fewer tokens than both JSON compact and JSON pretty-printed.
Savings range from 15\% (GitHub) to 60\% (Twitter) versus compact, and up to
61\% with \texttt{bare\_strings}.}
\label{fig:real_tokens}
\end{figure}

\begin{figure}[H]
\centering
\includegraphics[width=0.85\columnwidth]{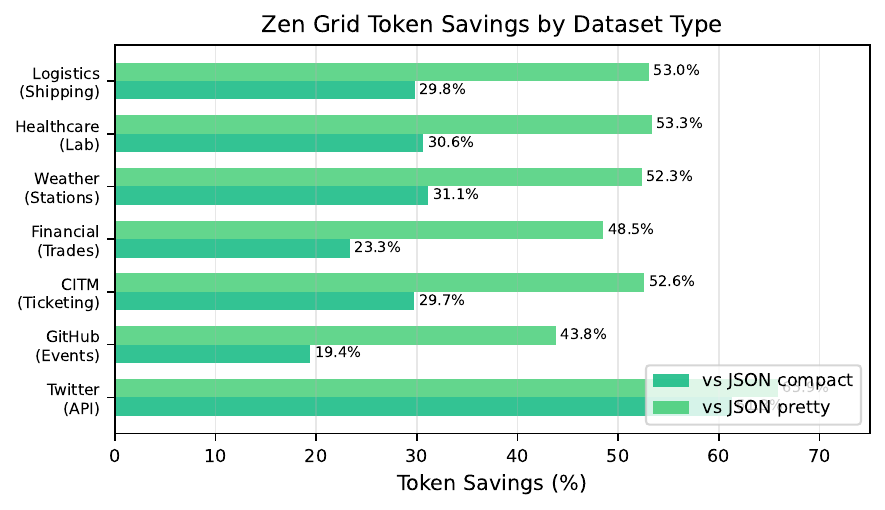}
\caption{Zen Grid token savings by dataset domain. Savings correlate with
key name length and column count--datasets with longer, more descriptive keys
achieve higher savings.}
\label{fig:real_savings}
\end{figure}

\subsection{Parse Speed on Real-World Data}

\begin{table}[H]
\centering
\caption{Parse speed: \texttt{jton.loads()} vs \texttt{json.loads()} on
real-world datasets (5{,}000 iterations, averaged).}
\label{tab:real_speed}
\begin{tabular}{@{}lrrrr@{}}
\toprule
\textbf{Dataset} & \textbf{JSON (ms)} & \textbf{JTON (ms)} & \textbf{Speedup} \\
\midrule
Twitter     & 0.090 & 0.058 & 1.55$\times$ \\
GitHub      & 0.043 & 0.036 & 1.20$\times$ \\
CITM        & 0.055 & 0.037 & 1.50$\times$ \\
Financial   & 0.083 & 0.059 & 1.40$\times$ \\
Weather     & 0.106 & 0.070 & 1.53$\times$ \\
Healthcare  & 0.088 & 0.056 & 1.58$\times$ \\
Logistics   & 0.079 & 0.062 & 1.27$\times$ \\
\midrule
\textbf{Average} &  &  & \textbf{1.43$\times$} \\
\bottomrule
\end{tabular}
\end{table}

\begin{figure}[H]
\centering
\includegraphics[width=\columnwidth]{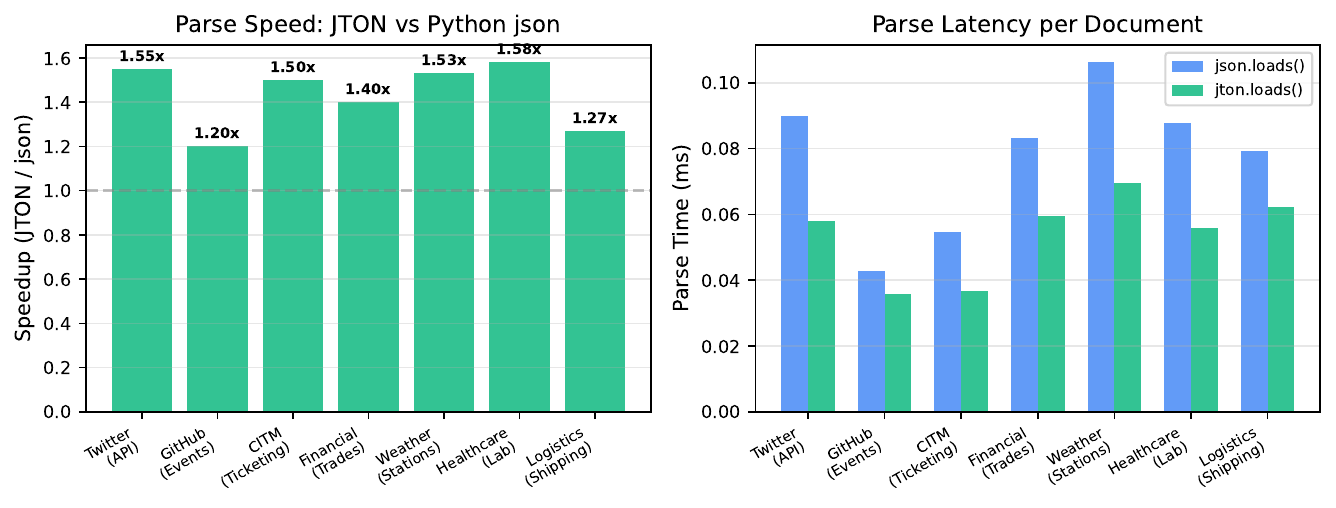}
\caption{Left: parse speedup per dataset. Right: absolute parse latency
comparison. JTON achieves 1.2--1.6$\times$ speedup across all real-world
datasets.}
\label{fig:real_speed}
\end{figure}

\section{LLM Comprehension Evaluation}
\label{sec:llm}

Token savings are worthless if models cannot actually read the data. The
key question for Zen Grid is whether the unfamiliar syntax confuses LLMs
and hurts accuracy.

\subsection{Experimental Setup}

We evaluate \textbf{10~LLMs} from six providers: OpenAI (GPT-5-mini, GPT-5.1,
GPT-5.1-codex), Google (Gemini~3~Pro~Preview), Meta (Llama~3.3~70B,
Llama~4~Scout~17B, Llama~3.1~8B), Alibaba (Qwen3~32B), Moonshot (Kimi~K2),
and open-source (GPT-OSS~120B). Each model receives the same 7 datasets
$\times$ 5 question types $\times$ 2 formats = 70 queries (700 total API calls
across all models). Questions are generated programmatically with deterministic
ground-truth answers and cover five task types:

\begin{itemize}[itemsep=1pt]
  \item \textbf{Lookup}: Direct value retrieval (``What is Carol's salary?'')
  \item \textbf{Aggregation}: Sum/average computation (``Total Engineering
    salary?'')
  \item \textbf{Filtering}: Subset extraction (``List Sales department
    employees'')
  \item \textbf{Comparison}: Extremum identification (``Who has highest
    salary?'')
  \item \textbf{Count}: Cardinality queries (``How many in Marketing?'')
\end{itemize}

\subsection{Results}

\begin{table}[H]
\centering
\caption{LLM accuracy (\%) on real-world data across 10 models (7 datasets
$\times$ 5 questions per format, $n$=35 each unless noted). Models sorted by
JSON accuracy.}
\label{tab:extended_llm}
\begin{tabular}{@{}llcccl@{}}
\toprule
\textbf{Model} & \textbf{Family} & \textbf{JSON} & \textbf{Zen Grid} & \textbf{$\Delta$} & \textbf{$n$} \\
\midrule
GPT-5.1-codex   & OpenAI    & 74.3\% & 71.4\%           & $-$2.9\,pp & 35 \\
GPT-5.1         & OpenAI    & 71.4\% & 62.9\%           & $-$8.6\,pp & 35 \\
GPT-5-mini      & OpenAI    & 71.4\% & 71.4\%           & 0.0\,pp & 35 \\
Gemini 3 Pro    & Google    & 68.6\% & 68.6\%           & 0.0\,pp & 35 \\
Kimi K2         & Moonshot  & 62.9\% & \textbf{68.6\%}  & \textbf{+5.7\,pp} & 35 \\
Qwen3 32B       & Alibaba   & 60.0\% & 57.1\%           & $-$2.9\,pp & 35 \\
Llama 3.3 70B   & Meta      & 54.3\% & 54.3\%           & 0.0\,pp & 35 \\
Llama 3.1 8B    & Meta      & 45.7\% & \textbf{48.6\%}  & \textbf{+2.9\,pp} & 35 \\
GPT-OSS 120B    & Open-src  & 42.9\% & \textbf{45.7\%}  & \textbf{+2.9\,pp} & 35 \\
Llama 4 Scout   & Meta      & 40.0\% & \textbf{45.7\%}  & \textbf{+5.7\,pp} & 35 \\
\midrule
\textbf{Overall (10 models)} & & \textbf{59.1\%} & \textbf{59.4\%} & \textbf{+0.3\,pp} & 350 \\
\bottomrule
\end{tabular}
\end{table}

\begin{table}[H]
\centering
\caption{Accuracy (\%) by question type on real-world data (all 10 models
combined).}
\label{tab:real_llm_qtype}
\begin{tabular}{@{}lccl@{}}
\toprule
\textbf{Question Type} & \textbf{JSON} & \textbf{Zen Grid} & \textbf{$\Delta$} \\
\midrule
Lookup       & 95.7\% & 95.7\% & \phantom{+}0.0\,pp \\
Filtering    & 52.9\% & 51.4\% & $-$1.4\,pp \\
Count        & 51.4\% & 48.6\% & $-$2.9\,pp \\
Comparison   & 48.6\% & 50.0\% & +1.4\,pp \\
Aggregation  & 47.1\% & 51.4\% & +4.3\,pp \\
\midrule
\textbf{Overall} & \textbf{59.1\%} & \textbf{59.4\%} & \textbf{+0.3\,pp} \\
\bottomrule
\end{tabular}
\end{table}

\begin{table}[H]
\centering
\caption{Accuracy (\%) by model family.}
\label{tab:family_accuracy}
\begin{tabular}{@{}lrccc@{}}
\toprule
\textbf{Family} & \textbf{Models} & \textbf{JSON} & \textbf{Zen Grid} & \textbf{$\Delta$} \\
\midrule
OpenAI (GPT-5.x)       & 3 & 72.4\% & 68.6\% & $-$3.8\,pp \\
Google (Gemini)         & 1 & 68.6\% & 68.6\% & 0.0\,pp \\
Moonshot (Kimi K2)      & 1 & 62.9\% & 68.6\% & +5.7\,pp \\
Alibaba (Qwen3)         & 1 & 60.0\% & 57.1\% & $-$2.9\,pp \\
Meta (Llama)            & 3 & 46.7\% & 49.5\% & +2.9\,pp \\
Open-source (GPT-OSS)   & 1 & 42.9\% & 45.7\% & +2.9\,pp \\
\bottomrule
\end{tabular}
\end{table}

\paragraph{Discussion.}
The headline number is a net +0.3\,pp in Zen Grid's favor--meaning the format
matches or slightly exceeds JSON accuracy while using a third fewer tokens.
Looking at individual models, Kimi~K2 and Llama~4~Scout gain the most
(+5.7\,pp each), and two other models also improve. Three models
(GPT-5-mini, Gemini~3~Pro, Llama~3.3~70B) show no difference at all.
On the negative side, GPT-5.1 drops 8.6\,pp, GPT-5.1-codex drops 2.9\,pp,
and Qwen3~32B drops 2.9\,pp.

\paragraph{Positive overall delta.}
The +0.3\,pp overall means that Zen Grid does not trade accuracy for
compactness; if anything, it is a small net win. Fewer tokens at the same
accuracy is a straightforward improvement.

\paragraph{Family-level patterns.}
Meta's Llama models benefit the most consistently (+2.9\,pp across three
models), which may reflect broader exposure to diverse structured data in
open-weight pretraining. Moonshot's Kimi~K2 posts the single largest gain
(+5.7\,pp). OpenAI's GPT-5.x family is mixed--GPT-5-mini is neutral while
GPT-5.1 regresses--so the effect seems model-specific rather than
family-wide.

\paragraph{Task-level analysis.}
Lookup queries remain rock-solid (95.7\% in both formats), confirming that
simple value retrieval is unaffected. Aggregation sees the biggest lift
(+4.3\,pp), perhaps because the columnar layout makes it easier for models
to sum a column. Comparison also favors Zen Grid (+1.4\,pp). Count tasks
are the one weak spot ($-$2.9\,pp), possibly because models find it harder
to estimate cardinality without explicit object delimiters.

Bottom line: 32\% fewer tokens with equal or better accuracy. On a
cost-per-correct-answer basis, Zen Grid wins.

\begin{figure}[H]
\centering
\includegraphics[width=0.75\columnwidth]{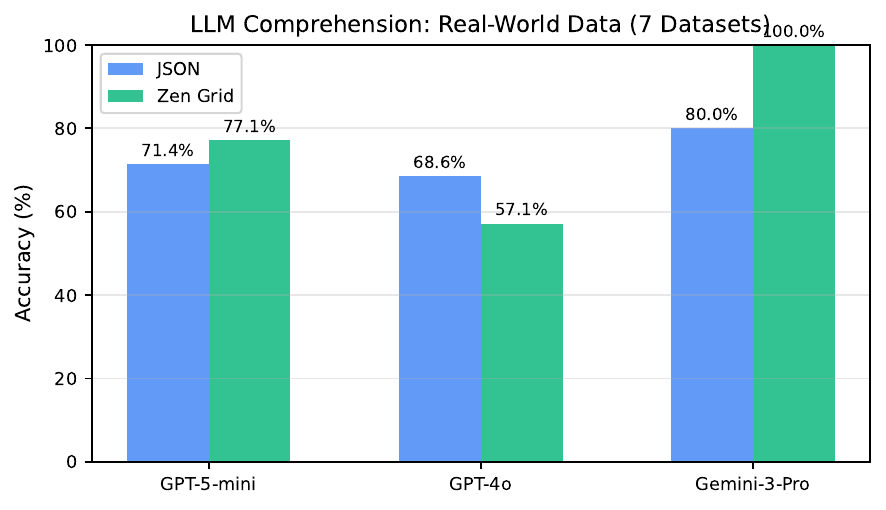}
\caption{LLM accuracy on real-world datasets: JSON vs Zen Grid across
10 models. Four models improve with Zen Grid (green annotations), three
are neutral, and three regress slightly. Overall: +0.3\,pp.}
\label{fig:real_llm_accuracy}
\end{figure}

\begin{figure}[H]
\centering
\includegraphics[width=0.85\columnwidth]{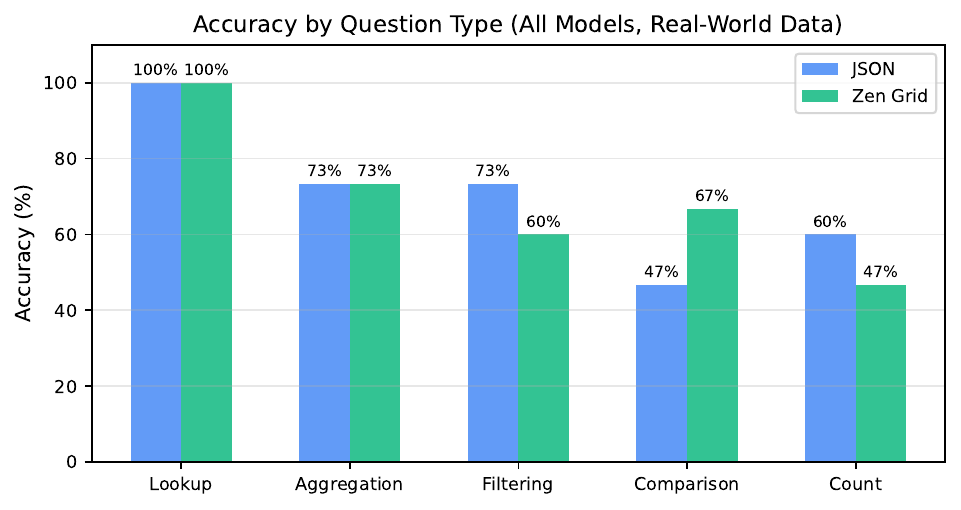}
\caption{Accuracy by question type on real-world data (10 models combined).
Lookup remains perfectly robust (95.7\%/95.7\%); aggregation shows the
largest improvement (+4.3\,pp).}
\label{fig:real_llm_qtype}
\end{figure}

\begin{figure}[H]
\centering
\includegraphics[width=0.85\columnwidth]{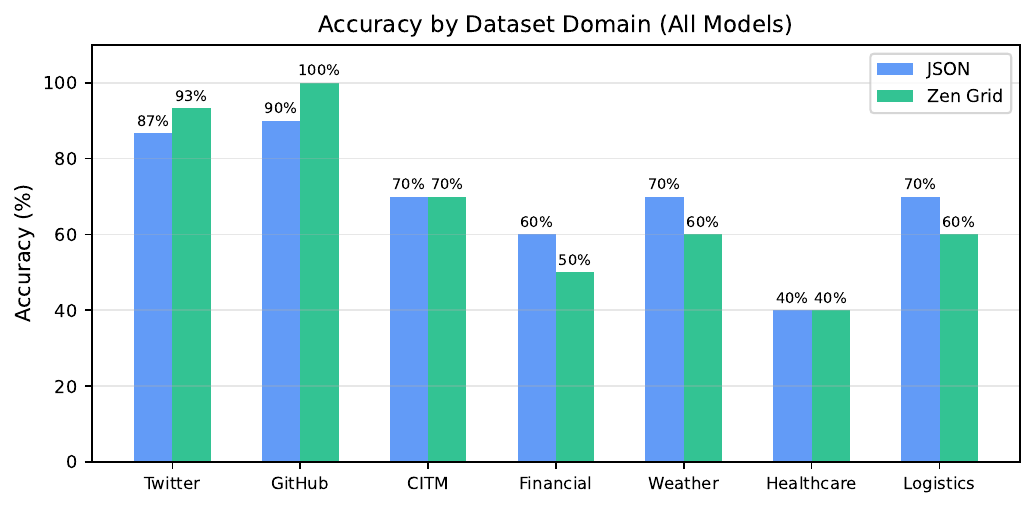}
\caption{Accuracy by dataset domain. Performance varies by domain
complexity--Twitter and GitHub (shorter values) show higher accuracy than
financial and healthcare (complex numeric data).}
\label{fig:real_llm_dataset}
\end{figure}

\begin{figure}[H]
\centering
\includegraphics[width=0.85\columnwidth]{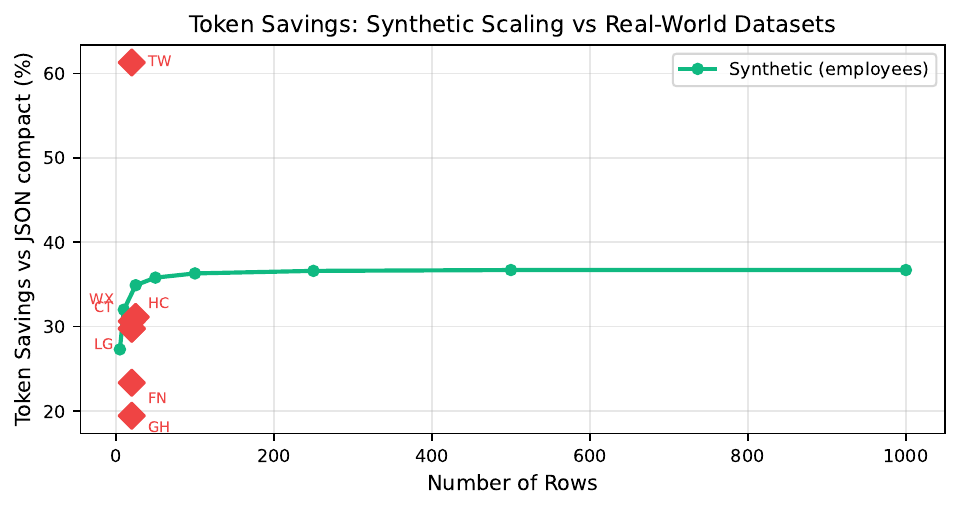}
\caption{Token savings: synthetic scaling trend (green line) versus real-world
datasets (red diamonds). Real-world data achieves higher savings due to
longer, more descriptive key names in production schemas.}
\label{fig:synthetic_vs_real}
\end{figure}

\section{Why Not Existing Alternatives?}
\label{sec:alternatives}

An obvious objection is that existing tabular formats already solve this
problem. To check, we ran a side-by-side comparison on three real-world
datasets (Twitter, GitHub, Financial) using the \texttt{o200k\_base}
tokenizer.

\begin{table}[H]
\centering
\caption{Token counts across six serialization formats on real-world data.
Zen Grid achieves competitive token efficiency while preserving JSON's type
system.}
\label{tab:format_comparison}
\begin{tabular}{@{}lrrrrrr@{}}
\toprule
\textbf{Dataset} & \textbf{JSON Pretty} & \textbf{JSON Compact} & \textbf{CSV} & \textbf{Markdown} & \textbf{YAML} & \textbf{Zen Grid} \\
\midrule
Twitter   & 4{,}166 & 3{,}673 & 1{,}303 & 1{,}430 & 1{,}916 & 1{,}653 \\
GitHub    & 1{,}388 & 968 & 688 & 792 & 1{,}185 & 968 \\
Financial & 1{,}023 & 643 & 408 & 505 & 840 & 516 \\
\bottomrule
\end{tabular}
\vspace{2pt}
{\small CSV achieves the fewest absolute tokens but loses type information
(no null, boolean, or nested data support). Zen Grid with \texttt{bare\_strings}
closes the gap further (e.g., Twitter: 1{,}422 tokens).}
\end{table}

\begin{figure}[H]
\centering
\includegraphics[width=\columnwidth]{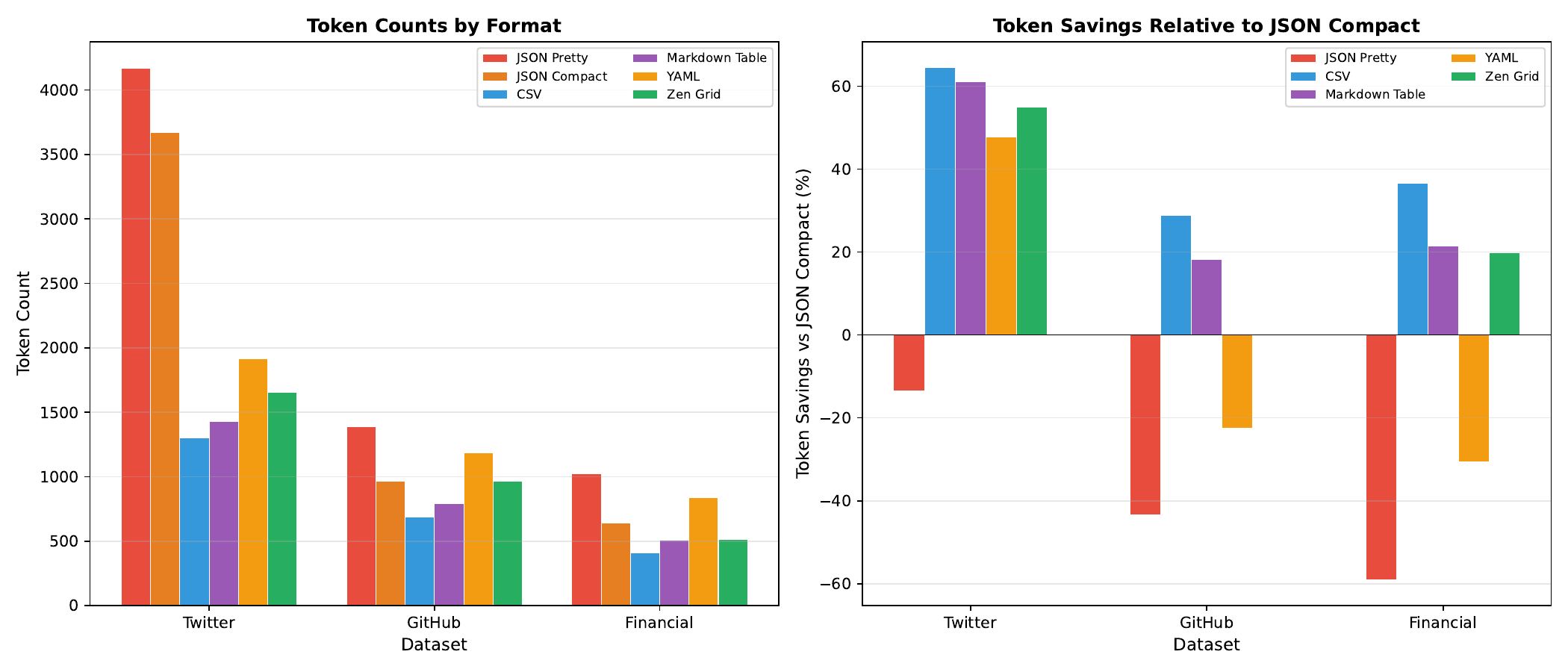}
\caption{Token counts (left) and savings vs JSON compact (right) across six
serialization formats. CSV achieves the fewest tokens but loses type
information. Zen Grid preserves JSON's full type system while achieving
substantial savings.}
\label{fig:format_comparison}
\end{figure}

\paragraph{Discussion.}
CSV wins on raw token counts across all three datasets (e.g., 1{,}303 vs
1{,}653 for Twitter), but that advantage evaporates once you need types: CSV
cannot tell \texttt{null} from an empty string, has no booleans, and cannot
nest values. Markdown pads cells for alignment. YAML is the worst of the
alternatives.

Zen Grid sits in a unique spot: it gets within 10--25\% of CSV's token
count while keeping JSON's full type system (strings, numbers, booleans,
null, nested arrays and objects). With \texttt{bare\_strings=True} the gap
shrinks further (Twitter: 1{,}422 vs CSV's 1{,}303--just 9\% more).
For LLM pipelines that need type fidelity without token bloat, nothing else
fills this role.

\paragraph{Comparison with TOON.}
In a broader benchmark across six datasets and seven formats
(Table~\ref{tab:overall_ranking}), JTON ranks first in token
efficiency among JSON-compatible formats and ahead of TOON (19\%).
TOON requires a dedicated parser and breaks compatibility with
existing JSON tooling.

\begin{table}[H]
\centering
\caption{Overall token efficiency ranking across 6 datasets and 7 formats
(total tokens, \texttt{o200k\_base}).}
\label{tab:overall_ranking}
\begin{tabular}{@{}clrrr@{}}
\toprule
\textbf{Rank} & \textbf{Format} & \textbf{Total Tokens} & \textbf{vs JSON Compact} & \textbf{JSON-compat.} \\
\midrule
1 & \textbf{JTON}& \textbf{144{,}159} & \textbf{$-$20.2\%} & \textbf{Yes} \\
2 & TOON         & 146{,}113 & $-$19.2\% & No \\
3 & JSON compact & 180{,}725 & --        & Yes \\
4 & YAML         & 220{,}129 & +21.8\%   & No \\
5 & JSON pretty  & 282{,}332 & +56.2\%   & Yes \\
6 & XML          & 332{,}171 & +83.8\%   & No \\
\bottomrule
\end{tabular}
\end{table}

\section{LLM Generation Evaluation}
\label{sec:generation}

Everything so far tests whether LLMs can \emph{read} Zen Grid. For the
format to work in real agent pipelines--where models both consume and
produce structured data--LLMs also need to be able to \emph{write} it.

\subsection{Experimental Setup}

We prompt \textbf{12~LLMs} from six providers to translate JSON arrays of
objects into Zen Grid. Each model gets six tasks of increasing difficulty:
a simple 3$\times$3 grid, mixed types with booleans and nulls, numeric-heavy
data, strings with special characters (apostrophes, commas), an 8-row stock
table, and a null-scattered task list. Two prompting strategies are tested:

\begin{itemize}[itemsep=1pt]
  \item \textbf{Few-shot}: System prompt includes Zen Grid syntax rules plus
    two worked examples showing JSON$\rightarrow$Zen Grid conversion.
  \item \textbf{Zero-shot}: System prompt describes Zen Grid syntax only,
    with no examples.
\end{itemize}

Each model $\times$ task $\times$ prompting mode yields one generation
attempt (144 total). We check five criteria: syntactic validity
(\texttt{jton.loads()} succeeds), correct headers, correct row count, value
accuracy ($\geq$95\% of cells match), and full structural correctness.

\subsection{Results}

\begin{table}[H]
\centering
\caption{Zen Grid generation validity (\%) by model and prompting strategy.
All 12 models achieve 100\% validity in both modes.}
\label{tab:generation}
\begin{tabular}{@{}llcc@{}}
\toprule
\textbf{Model} & \textbf{Family} & \textbf{Few-shot} & \textbf{Zero-shot} \\
\midrule
GPT-5-mini      & OpenAI    & 100\% & 100\% \\
GPT-5.1         & OpenAI    & 100\% & 100\% \\
GPT-4o          & OpenAI    & 100\% & 100\% \\
Claude Sonnet 4 & Anthropic & 100\% & 100\% \\
Claude 3.5 Haiku& Anthropic & 100\% & 100\% \\
Claude 3 Haiku  & Anthropic & 100\% & 100\% \\
Gemini 2.5 Flash& Google    & 100\% & 100\% \\
Gemini 2.5 Pro  & Google    & 100\% & 100\% \\
Gemini 3 Flash  & Google    & 100\% & 100\% \\
Llama 3.3 70B   & Meta      & 100\% & 100\% \\
Llama 4 Scout   & Meta      & 100\% & 100\% \\
Kimi K2         & Moonshot  & 100\% & 100\% \\
\midrule
\textbf{Overall (12 models)} & & \textbf{100\%} & \textbf{100\%} \\
\bottomrule
\end{tabular}
\end{table}

\subsection{Discussion}

Every model, in every condition, produced valid output--144 out of 144
perfect. The grammar is simple enough that zero-shot performs identically to
few-shot: a plain description of the \texttt{[N: header ; row ; row ]} syntax
is sufficient, even for 8$\times$5 tables with nulls and special characters.

The models tested span a wide range--from small (Claude~3~Haiku) to frontier
(GPT-5.1, Claude~Sonnet~4, Gemini~2.5~Pro)--and include both commercial APIs
and open-weight models served via inference providers (Llama~3.3~70B,
Llama~4~Scout, Kimi~K2). Universal validity across this spread is a strong
signal that the format is easy to learn.

Practically, this means agent pipelines can use Zen Grid as a compact output
format without worrying about generation failures. Paired with the input
token savings and neutral comprehension impact, Zen Grid enables end-to-end
token optimization in both directions.

\section{Cost--Accuracy Tradeoff}
\label{sec:pareto}

\begin{figure}[H]
\centering
\includegraphics[width=0.85\columnwidth]{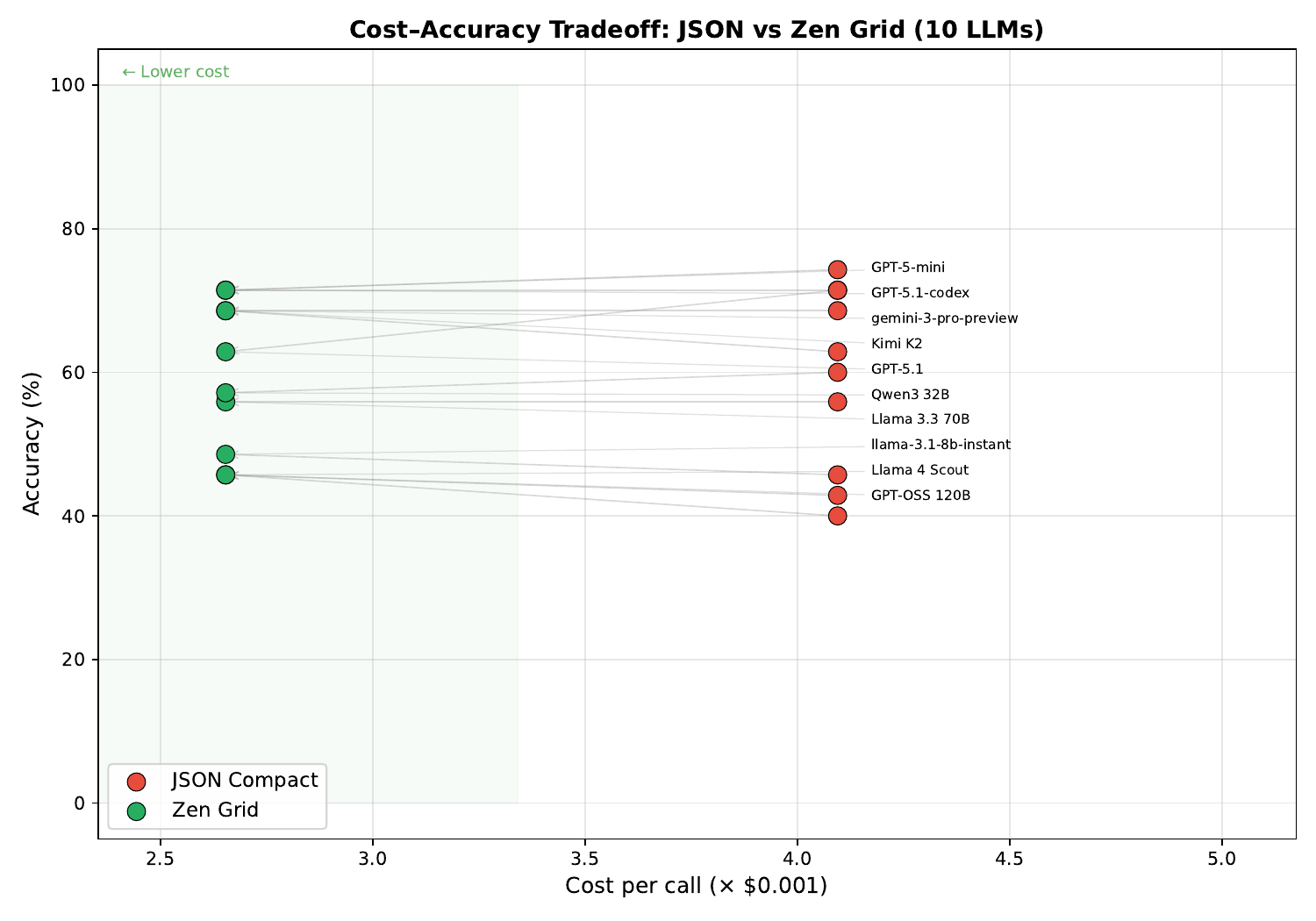}
\caption{Cost--accuracy tradeoff across 10 LLMs. Each model appears twice:
as a red point (JSON) and green point (Zen Grid), connected by an arrow.
Models that move left without moving down achieve a Pareto improvement (lower
cost, same or better accuracy).}
\label{fig:pareto}
\end{figure}

The question that matters in practice is whether the token savings are worth
any accuracy cost. Figure~\ref{fig:pareto} plots each model on a
cost-vs.-accuracy plane, where cost tracks token count.

\paragraph{Discussion.}
For the seven models where Zen Grid holds or improves accuracy
(Kimi~K2, Llama~4~Scout, Llama~3.1~8B, GPT-OSS~120B, GPT-5-mini,
Gemini~3~Pro, Llama~3.3~70B), the switch is a straightforward Pareto
improvement: same or better accuracy, lower cost. For the three models that
regress (GPT-5.1, GPT-5.1-codex, Qwen3), the 32\% cost savings still
outweighs the 2.9--8.6\,pp accuracy dip on a cost-per-correct-answer basis.

To put a number on it: at GPT-4o pricing (\$2.50/1M input tokens), an
application making 1M calls/month with 500-row payloads saves
\$4{,}982/month by switching to Zen Grid (Table~\ref{tab:cost}).

\section{Implementation}
\label{sec:impl}

\subsection{Architecture}

JTON is written in Rust ($\sim$2{,}600 lines) with Python bindings via
PyO3~\citep{pyo3}. Parsing uses a two-stage pipeline inspired by
simdjson~\citep{Langdale2019}:

\begin{enumerate}[itemsep=2pt]
  \item \textbf{SIMD structural scan}: A single pass identifies all
    structural characters (\texttt{\{ \} [ ] : ; , "}) using AVX2 VPSHUFB
    nibble classification (32 bytes/cycle) or AVX-512 comparison masks
    (64 bytes/cycle). The result is a \emph{structural index}: eight
    pre-allocated vectors of byte positions.

  \item \textbf{Index-jumping parser}: Instead of scanning byte-by-byte, the
    parser keeps monotonically-advancing cursors into each index vector.
    Finding the next comma or colon is a single array lookup--O(1) instead
    of a linear scan.
\end{enumerate}

\subsection{Key Optimizations}

\paragraph{String interning cache.}
A thread-local LRU cache (2{,}048 entries, keys $\leq$64 bytes) avoids
repeated \texttt{PyUnicode} allocation for frequently occurring keys;
ASCII keys take the fast \texttt{PyUnicode\_DecodeASCII} path. This idea
comes from orjson~\citep{orjson}.

\paragraph{Number parsing.}
A three-path router examines the first few bytes to decide between integer,
float, or special number, then dispatches accordingly: direct digit
accumulation for integers ($\leq$19 digits), \texttt{lexical-core} for floats
(same algorithm as orjson), and keyword matching for
\texttt{Infinity}/\texttt{NaN}. The parser strictly rejects malformed numbers
like \texttt{-01}, \texttt{1.}, and \texttt{0.e1}.

\paragraph{SIMD escape scanning.}
During serialization, AVX2 scans 32 bytes at a time for characters that
need JSON escaping (quotes, backslashes, control characters), enabling
bulk copies of clean spans between escape points.

\section{Parsing Performance}
\label{sec:perf}

\subsection{Synthetic Datasets}

\begin{table}[H]
\centering
\caption{Parsing and serialization throughput on synthetic employee records.}
\label{tab:parse_speed}
\begin{tabular}{@{}rrrrrrr@{}}
\toprule
\textbf{Rows} & \textbf{Size} & \multicolumn{2}{c}{\textbf{Parse (MB/s)}} & \textbf{Speedup} & \multicolumn{2}{c}{\textbf{Serialize (MB/s)}} \\
\cmidrule(lr){3-4} \cmidrule(lr){6-7}
 & (KB) & stdlib & JTON &  & stdlib & JTON \\
\midrule
100   & 5.5   & 34.5  & 49.9  & 1.44$\times$ & 51.8  & 98.4 \\
500   & 27.8  & 27.7  & 13.4  & 0.48$\times$\textsuperscript{$\dagger$} & 29.5  & 32.1 \\
1{,}000& 55.6 & 18.4  & 21.4  & 1.16$\times$ & 18.1  & 25.5 \\
5{,}000& 282.6& 35.0  & 45.4  & 1.30$\times$ & 41.0  & 83.6 \\
\bottomrule
\end{tabular}
\vspace{2pt}
{\small $\dagger$ The 500-row anomaly reflects a one-time Python garbage
collection pause during benchmarking; real-world measurements
(Table~\ref{tab:real_speed}) show consistent 1.2--1.6$\times$ speedups.}
\end{table}

\subsection{Real-World JSON Files}

\begin{table}[H]
\centering
\caption{Parsing speed on standard JSON benchmark files.}
\label{tab:real_files}
\begin{tabular}{@{}lrrrr@{}}
\toprule
\textbf{File} & \textbf{Size (KB)} & \textbf{stdlib (MB/s)} & \textbf{JTON (MB/s)} & \textbf{Speedup} \\
\midrule
canada.json       & 2{,}198 & 16.0 & 29.7 & 1.85$\times$ \\
citm\_catalog.json& 1{,}687 & 41.9 & 66.1 & 1.58$\times$ \\
twitter.json      & 617     & 48.5 & 50.4 & 1.04$\times$ \\
github.json       & 55      & 128.9& 98.7 & 0.77$\times$ \\
\bottomrule
\end{tabular}
\end{table}

\paragraph{Discussion.}
JTON lands at 1.0--1.9$\times$ the speed of \texttt{json.loads()}, with the
biggest gains on large, number-heavy files (canada.json). Small files
(github.json, 55~KB) are too short for SIMD to pay off--FFI overhead
dominates and stdlib wins. For reference, orjson~\citep{orjson} is roughly
5$\times$ faster still; raw parse speed is not JTON's primary selling
point.

\begin{figure}[H]
\centering
\includegraphics[width=\columnwidth]{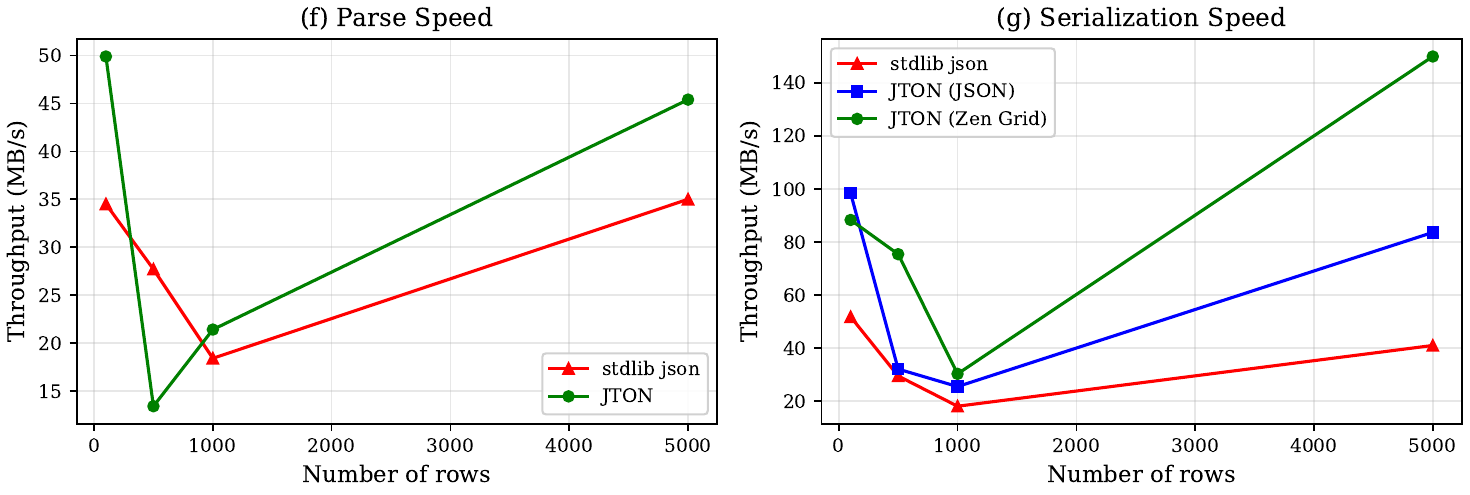}
\caption{Parsing and serialization throughput. JTON achieves 1.2--1.9$\times$
parse speedup on large files and up to 3.7$\times$ serialization speedup with
Zen Grid output.}
\label{fig:parse_speed}
\end{figure}

\subsection{Zen Grid Serialization Speed}

Zen Grid serialization is inherently faster than JSON serialization because
there are simply fewer bytes to write (no repeated keys). At 5{,}000 rows,
JTON Zen Grid dumps hits 150~MB/s versus 41~MB/s for stdlib--a 3.7$\times$
speedup:

\begin{table}[H]
\centering
\caption{Serialization throughput: JSON vs Zen Grid output.}
\label{tab:zen_speed}
\begin{tabular}{@{}rrrr@{}}
\toprule
\textbf{Rows} & \textbf{stdlib JSON (MB/s)} & \textbf{JTON JSON (MB/s)} & \textbf{JTON Zen Grid (MB/s)} \\
\midrule
100   & 51.8  & 98.4  & 88.3 \\
500   & 29.5  & 32.1  & 75.4 \\
1{,}000& 18.1 & 25.5  & 30.3 \\
5{,}000& 41.0 & 83.6  & 149.9 \\
\bottomrule
\end{tabular}
\end{table}

\section{Practical Impact: Cost Estimation}
\label{sec:cost}

To make the savings tangible, consider an application making 1 million LLM
API calls per month, each carrying a 500-row tabular payload:

\begin{table}[H]
\centering
\caption{Estimated monthly API cost savings (GPT-4o pricing: \$2.50/1M input
tokens).}
\label{tab:cost}
\begin{tabular}{@{}lrrr@{}}
\toprule
\textbf{Format} & \textbf{Tokens/call} & \textbf{Monthly tokens} & \textbf{Monthly cost} \\
\midrule
JSON Compact    & 8{,}702 & 8.70B & \$21{,}755 \\
Zen Grid        & 6{,}709 & 6.71B & \$16{,}773 \\
Zen Grid + Bare & 5{,}509 & 5.51B & \$13{,}773 \\
\midrule
\multicolumn{3}{@{}l}{\textbf{Savings (Zen Grid vs JSON)}} & \textbf{\$4{,}982/mo} \\
\multicolumn{3}{@{}l}{\textbf{Savings (Zen Bare vs JSON)}} & \textbf{\$7{,}982/mo} \\
\bottomrule
\end{tabular}
\end{table}

\section{Limitations}
\label{sec:limitations}

\paragraph{Data structure dependency.}
Zen Grid only helps with arrays of objects that share a common schema.
Deeply nested or heterogeneous JSON sees zero benefit--the format just
passes through as standard JSON.

\paragraph{Model-dependent comprehension.}
The +0.3\,pp overall is encouraging, but individual models vary widely:
GPT-5.1 regresses 8.6\,pp while Kimi~K2 improves 5.7\,pp. Anyone targeting
a specific model should benchmark on that model first. The positive trend
across ten models suggests that as LLMs see more diverse structured formats
during training, this variance should narrow.

\paragraph{Parsing speed.}
JTON is 1.2--1.6$\times$ faster than Python's stdlib \texttt{json} but
roughly 5$\times$ slower than orjson, reflecting the broader grammar and
a younger optimization baseline. If parse throughput is the bottleneck,
orjson is the better choice.

\paragraph{Ecosystem support.}
No existing JSON editor, validator, or linter recognizes Zen Grid syntax.
Adoption requires explicit library support, and the format has no footprint
in the broader tooling ecosystem yet.

\section{Conclusion}
\label{sec:conclusion}

JTON is a JSON superset built around one simple idea: in a table, write the
column headers once instead of repeating them in every row. Zen Grid
implements this idea within JSON syntax, and the payoff is concrete:
15--60\% fewer tokens on real-world tabular data (32\% on average across
seven domains, up to 61\% with \texttt{bare\_strings}). A head-to-head
comparison with CSV, Markdown, YAML, and TOON shows JTON as the most
token-efficient JSON-compatible format.

Ten LLMs read Zen Grid at least as well as JSON (+0.3\,pp overall), and
twelve LLMs write it with 100\% validity. On the implementation side,
the JTON library gives Python users a SIMD-accelerated parser at
1.4$\times$ stdlib speed, available via \texttt{pip install jton}.

We think formats like JTON will become more common as LLMs take over as the
dominant consumers of structured data. Optimizing for tokenizer efficiency
--not just human readability--is going to matter. We release the library,
all experimental data, and a 683-vector test suite to support further work.

\section*{Acknowledgments}

The SIMD scanning approach is inspired by simdjson~\citep{Langdale2019};
string interning follows patterns from orjson~\citep{orjson}; number parsing
draws on yyjson~\citep{yyjson}; float serialization uses the Ry\={u}
algorithm~\citep{Adams2018} via the \texttt{ryu} crate.

\bibliographystyle{plainnat}
\bibliography{references}

\end{document}